\documentclass[10pt,twocolumn,letterpaper]{article}

\usepackage[pagenumbers]{cvpr} 

\usepackage[dvipsnames]{xcolor}

\definecolor{cvprblue}{rgb}{0.21,0.49,0.74}
\usepackage[pagebackref,breaklinks,colorlinks,citecolor=cvprblue]{hyperref}

\def\paperID{1738} 
\def\confName{CVPR}
\def\confYear{2024}

\def\pp{EgoTAP}
\title{Attention-Propagation Network for Egocentric Heatmap to 3D Pose Lifting}

\author{Taeho Kang \\
Seoul National University, South Korea \\
{\tt\small taeho.kang@hcs.snu.ac.kr}
\and
Youngki Lee \\
Seoul National University, South Korea \\
{\tt\small youngkilee@snu.ac.kr}
}

\begin{document}

\maketitle
\begin{abstract}
We present \pp{}, a heatmap-to-3D pose lifting method for highly accurate stereo egocentric 3D pose estimation. Severe self-occlusion and out-of-view limbs in egocentric camera views make accurate pose estimation a challenging problem. To address the challenge, prior methods employ joint heatmaps-probabilistic 2D representations of the body pose, but heatmap-to-3D pose conversion still remains an inaccurate process. We propose a novel heatmap-to-3D lifting method composed of the Grid ViT Encoder and the Propagation Network. The Grid ViT Encoder summarizes joint heatmaps into effective feature embedding using self-attention. Then, the Propagation Network estimates the 3D pose by utilizing skeletal information to better estimate the position of obscure joints. Our method significantly outperforms the previous state-of-the-art qualitatively and quantitatively demonstrated by a 23.9\% reduction of error in an MPJPE metric. Our source code is available in GitHub~\footnote{\url{https://github.com/tho-kn/EgoTAP}}.
\end{abstract}
\section{Introduction}
\label{sec:intro}
\begin{figure}[pt]
    \centering
    \begin{subfigure}{0.9\linewidth}
        \centering
        \includegraphics[width=\linewidth]{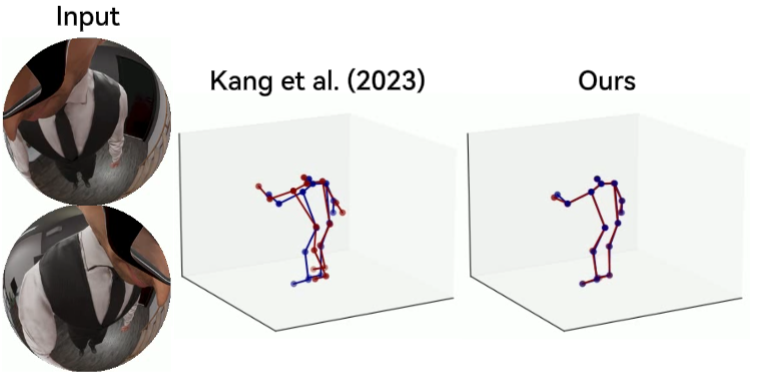}
    \end{subfigure}
    \caption{The stereo egocentric input and the comparison of the estimated pose of the state-of-the-art method~\cite{kang2023ego3dpose} and ours. Blue color for the ground truth and red color for the respective method's estimation}\label{fig:intro_comparison}
    \vspace{-10pt}
\end{figure}

The increasing use of Virtual Reality(VR) and Augmented Reality(AR) applications has prompted efforts to perform various vision tasks with minimal wearable sensors. Specifically, head-mounted cameras in the egocentric setup (Fig.~\ref{fig:intro_comparison}) received increasing attention thanks to their accessibility. Here, accurate 3D pose estimation is noted as a task critical for seamlessly integrating virtual selves into the real world. However, existing egocentric pose estimation methods still suffer from accuracy challenges~\cite{kang2023ego3dpose}.
 
\begin{figure}[pt]
    \centering
    \begin{subfigure}{0.8\linewidth}
        \centering
        \includegraphics[width=\linewidth]{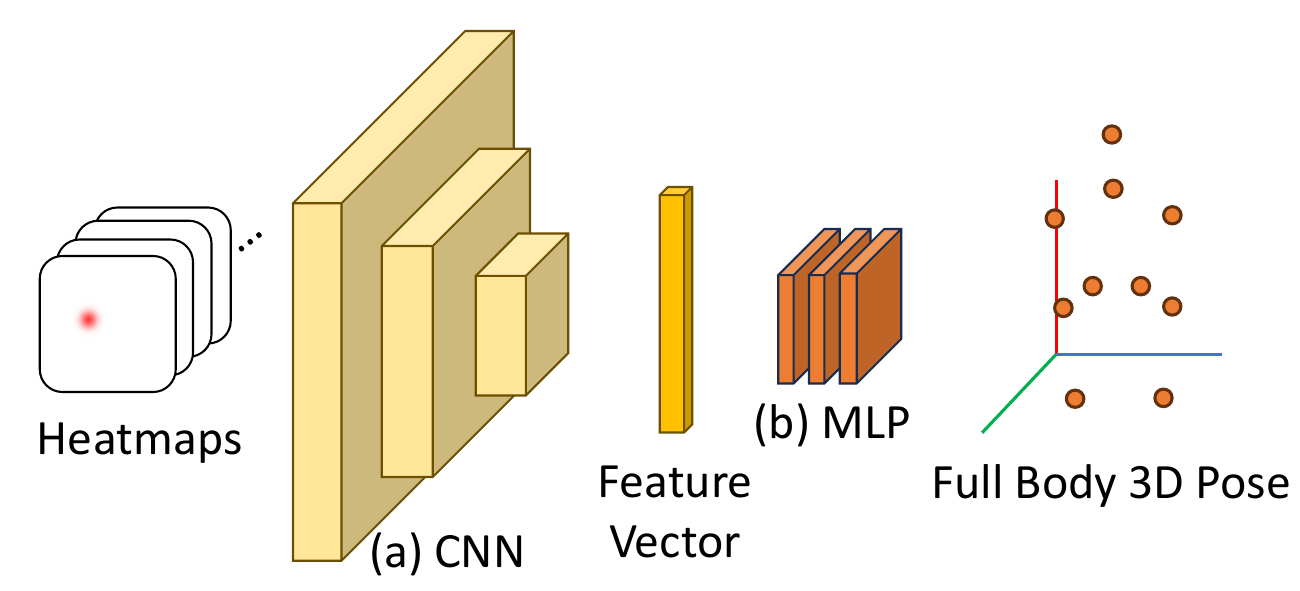}
    \end{subfigure}
    \caption{The architecture of the common baseline heatmap-to-3D approach. This architecture is adopted by monocular $x$R-EgoPose~\cite{tome2019xr} and stereo UnrealEgo~\cite{hakada2022unrealego} for 3D pose inference.}
    \label{fig:lifting_baseline}
    \vspace{-10pt}
\end{figure}

Conventional 3D pose estimation methods typically derive 3D pose directly from 2D pose information \cite{shan2023diffusion, li2023multi, zheng20213d}. However, this approach faces challenges in egocentric setups due to inaccuracies in 2D pose estimation resulting from limited camera views and self-occlusion. To address this, egocentric pose estimation methods use joint heatmaps—probabilistic 2D representations of joints \cite{conf/cvpr/TompsonGJLB15}. These heatmaps employ probability distributions of likely joint positions rather than exact locations. Following this approach, methods generate heatmaps for key joints from egocentric camera input, consolidate them into a unified feature embedding vector, and perform full-body 3D pose estimation (Fig.~\ref{fig:lifting_baseline}). However, two critical problems in the heatmap-to-3D lifting process significantly impact position estimation accuracy.

\begin{figure}[pt]
    \centering
    \begin{subfigure}{0.45\linewidth}
        \centering
        \begin{subfigure}{0.4\linewidth}
            \centering
            \includegraphics[width=\linewidth]{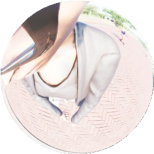}
        \end{subfigure}%
        \hspace{0.05\linewidth}
        \begin{subfigure}{0.4\linewidth}
            \centering
            \includegraphics[width=\linewidth]{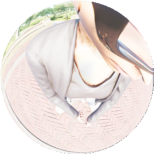}
        \end{subfigure}
        \caption{Input}
    \end{subfigure}
    \hspace{0.05\linewidth}
    \begin{subfigure}{0.45\linewidth}
        \centering
        \begin{subfigure}{0.4\linewidth}
        \includegraphics[width=\linewidth]{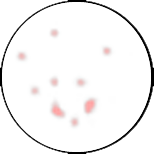}
        \end{subfigure}%
        \hspace{0.05\linewidth}
        \begin{subfigure}{0.4\linewidth}
            \includegraphics[width=\linewidth]{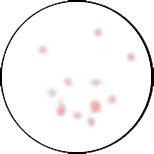}
        \end{subfigure}%
        \caption{Estimated Heatmaps}
    \end{subfigure}
    \\
    \begin{subfigure}{0.45\linewidth}
        \centering
        \begin{subfigure}{0.4\linewidth}
            \includegraphics[width=\linewidth]{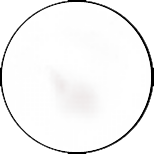}
        \end{subfigure}%
        \hspace{0.05\linewidth}
        \begin{subfigure}{0.4\linewidth}
            \includegraphics[width=\linewidth]{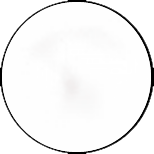}
        \end{subfigure}
        \caption{CNN Encoder}
    \end{subfigure}
    \hspace{0.05\linewidth}
    \begin{subfigure}{0.45\linewidth}
        \centering
        \begin{subfigure}{0.4\linewidth}
            \includegraphics[width=\linewidth]{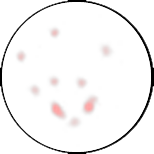}
        \end{subfigure}%
        \hspace{0.05\linewidth}
        \begin{subfigure}{0.4\linewidth}
            \includegraphics[width=\linewidth]{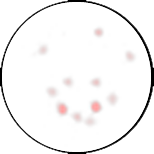}
        \end{subfigure}%
        \caption{Grid ViT Encoder}
    \end{subfigure}
    \\
    \caption{Comparison of the reconstructed heatmaps from the encoded heatmap features, with the frozen encoder from (c) CNN Encoder and (d) Grid ViT Encoder of the pose estimation model.}\label{fig:heatmap_reconstruction}
\end{figure}

\textbf{Inefficiency in feature embedding.} 
Obtaining an effective feature embedding from the heatmap poses a significant challenge. A robust embedding vector is crucial for accurately reconstructing the 3D pose, given the indirect mapping between the probabilistic, high-dimensional heatmaps and the 3D pose. However, the standard design, utilizing a CNN (Convolutional Neural Network) encoder, proves inadequate for feature summarization. The CNN encoder fails to preserve correspondence between specific heatmaps and joint poses, as features are merged into a single shared embedding. Furthermore, the spatial locality assumption of CNNs does not hold in an egocentric setup, where related joints may be distant in pixel space due to the proximity of ego-centric cameras to body parts and biased positions. The 3D pose lifting employs heatmap reconstruction loss~\cite{tome2019xr, zhao2021egoglass, hakada2022unrealego, kang2023ego3dpose} to recover heatmap information, but full recovery becomes challenging once the embedding vector has significantly lost information, as illustrated in Fig.~\ref{fig:heatmap_reconstruction}.

\textbf{Feature Importance-agnostic 3D Lifting.} 
Secondly, there is a significant inaccuracy in estimating a full-body 3D pose without effectively distinguishing between important and unimportant features, as seen in the conventional pipeline using Multi-Layer Perception (Fig.~\ref{fig:lifting_baseline} (b)). The prior methods \cite{tome2019xr, zhao2021egoglass, hakada2022unrealego, kang2023ego3dpose} do not consider the certainty of joints or the physical relationships between them, relying solely on the motion distribution within the training data. This approach may result in obscure joint features adversely affecting joints with clear visual cues in the camera or those estimable from nearby joint information. The supplementary material highlights that body extremities with less visibility exhibit higher estimation errors.

To tackle these challenges, we introduce \pp{} (Egocentric Transformer-Attention Propagation Network). \pp{} incorporates two key techniques: Grid ViT (Vision Transformer) Heatmap Encoder and Propagation Network. We design the former to generate an effective feature embedding that (i) preserves the correspondence between heatmaps and feature embedding and (ii) captures meaningful relationships between distant pixels. The latter assigns weights to evident joint features with clearer visual cues and predicts the position of less visible joints using the skeletal information of body limbs. Through these techniques, we achieve a substantial improvement in pose error metrics, demonstrating a 23.9\% reduction in MPJPE and a 17.7\% decrease in PA-MPJPE compared to state-of-the-art methods. 

\textbf{Grid ViT Heatmap Encoder} addresses the inefficiency of the CNN encoding process. The Grid ViT Heatmap Encoder consolidates all joint heatmaps into a single image and divides them into patches, with each patch corresponding to a heatmap. Subsequently, self-attention is applied across all patches, generating per-patch feature embeddings. The ViT Heatmap Encoder offers two key advantages. Firstly, the per-patch embedding better preserves the position information of the original joint heatmaps. Secondly, self-attention facilitates the effective embedding of inter-joint relationships, particularly useful for joint features in distant areas.

\textbf{Propagation Network} propagates various features from the neck joint, likely to have the evident features, to the body's extremities with less visibility, following the body hierarchy. To enable propagation, we devise an LSTM~\cite{LSTM1997}-inspired cell, PU (Propagation Unit). The PU takes the parent joint's feature, the relational (limb) features as a hidden state, and the child joint's features as input to predict the final 3D position. The PU has an additional gate to forget the parent and relational features in case the child joint features are evident, limiting the role of the predictive estimation only for obscure joints. This design explicitly leverages the physical relationships of joints rather than implicitly inferring them from the training data, thereby contributing to higher pose estimation accuracy.

In summary, our contributions are the following:
\begin{itemize}
\item{The first egocentric 3D pose estimation method using a vision transformer for efficient feature embedding.}
\item{The Propagation Network that enables the predictive estimation for obscure joints using skeletal hierarchy.}
\item{The Propagation Unit, to control the importance of the propagated features.}
\item{\pp{} outperforms the state-of-the-art stereo egocentric pose estimation both qualitatively and quantitatively.}
\end{itemize}

\section{Related Works}
\subsection{Egocentric Pose Estimation}
Egocentric pose estimation can be classified into two main categories. The first category focuses on estimating the pose of other people within the camera's field of view, as in Ng et al.\cite{ng2019you2me} while the second category estimates the pose of the user self~\cite{li2023ego}. Our work belongs to the second category, especially with a downward-oriented egocentric camera.

EgoCap~\cite{rhodin2016egocap} showcased its potential using stereo cameras on a helmet-mounted stick. 
Mo$^2$Cap$^2$~\cite{xu2019mo2cap2} and $x$R-EgoPose~\cite{tome2019xr} have introduced single-camera methods, which handle occlusion. The former proposes a two-branched heatmap, one for the lower body with a magnified view. The latter adds a heatmap reconstructor to preserve the probabilistic information of heatmaps. Recent methods utilize an external camera view to make a weakly labeled large-scale dataset~\cite{wang2022estimating} and a scene depth estimation model to estimate 3D pose with volumetric heatmaps~\cite{wang2023scene}. These methods, however, require additional external cameras or depth datasets from specific views.

Recently, a stereo egocentric setup has gained attention for a wide-view stereo perspective. EgoGlass~\cite{zhao2021egoglass} introduces an unobtrusive eyeglass-mounted stereo camera setup, minimizing obtrusiveness. It incorporates an additional segmentation branch on the heatmap estimator module to improve the awareness of body parts and pixel correspondence. UnrealEgo~\cite{hakada2022unrealego} introduces a publicly available synthetic large-scale dataset based on the EgoGlass setup and proposes to share weights and merge features across the stereo view in the heatmap estimator. Ego3DPose~\cite{kang2023ego3dpose} suggests making an independent estimate of the 3D orientation of each limb, using the concatenated orientation vector for the final decoder. We observed two problems in these prior works, i.e., information loss in feature embedding and data-dependant estimation of obscure joints, and propose two corresponding techniques to address the problems. 

\subsection{3D Human Pose Estimation with Transformer}
The transformer-based architecture has been explored for the 3D pose estimation task. Epipolar Transformers~\cite{Epipolar2020CVPR} utilizes attention to match features along the epipolar line from the stereo view. Most methods focused on using transformers for 2D to 3D pose lifting spatially and temporally. PoseFormer~\cite{zheng20213d} is the first transformer-based 2D-to-3D pose lifting method consisting of spatial and temporal transformer networks. MixSTE~\cite{Zhang_2022_CVPR} and PoseFormerV2~\cite{Zhao_2023_CVPR} improved it with the per joint temporal characteristics and frequency domain feature. Unlike prior works, we exploit the transformer to effectively embed heatmap information for accurate heatmap-to-3D pose lifting.

\subsection{Skeletal Network Models}
Multiple works utilize skeletal hierarchy for vision tasks. For instance, Liu et al.~\cite{10.1109/TPAMI.2017.2771306} uses spatio-temporal LSTM to iterate through all joints for action recognition. Most recent efforts utilize a graph-based model to represent skeletal hierarchy. The Graph Convolutional Networks~\cite{DBLP:conf/iclr/KipfW17} is widely utilized for activity recognition~\cite{GCN2022Action} while ST-GCN~\cite{Yan_Xiong_Lin_2018} models a dynamic skeletal graph in a spatiotemporal manner. The graph-based models are adapted for the pose estimation~\cite{GCN2021Pose, Yu_2023_ICCV, Yan_Xiong_Lin_2018}, using dynamic skeletal graphs with action-specific edges or adopting adaptive ST-GCN~\cite{Yu_2023_ICCV, Yan_Xiong_Lin_2018}.

Our work is the first to leverage skeletal information in the ego-centric setup. Specifically, we address the challenge of obscure features, particularly for body extremities, which impact the pose estimation of all body parts. Introducing a skeleton-aware uni-directional Propagation Network model, we leverage clear visual cues from camera-proximate joints to estimate the pose of body parts with obscure visual features.

\section{Method}
\subsection{Overview}
\label{sec:overview}
\begin{figure*}[pt]
    \centering
    \begin{subfigure}{1.0\linewidth}
        \centering
        \includegraphics[width=\linewidth]{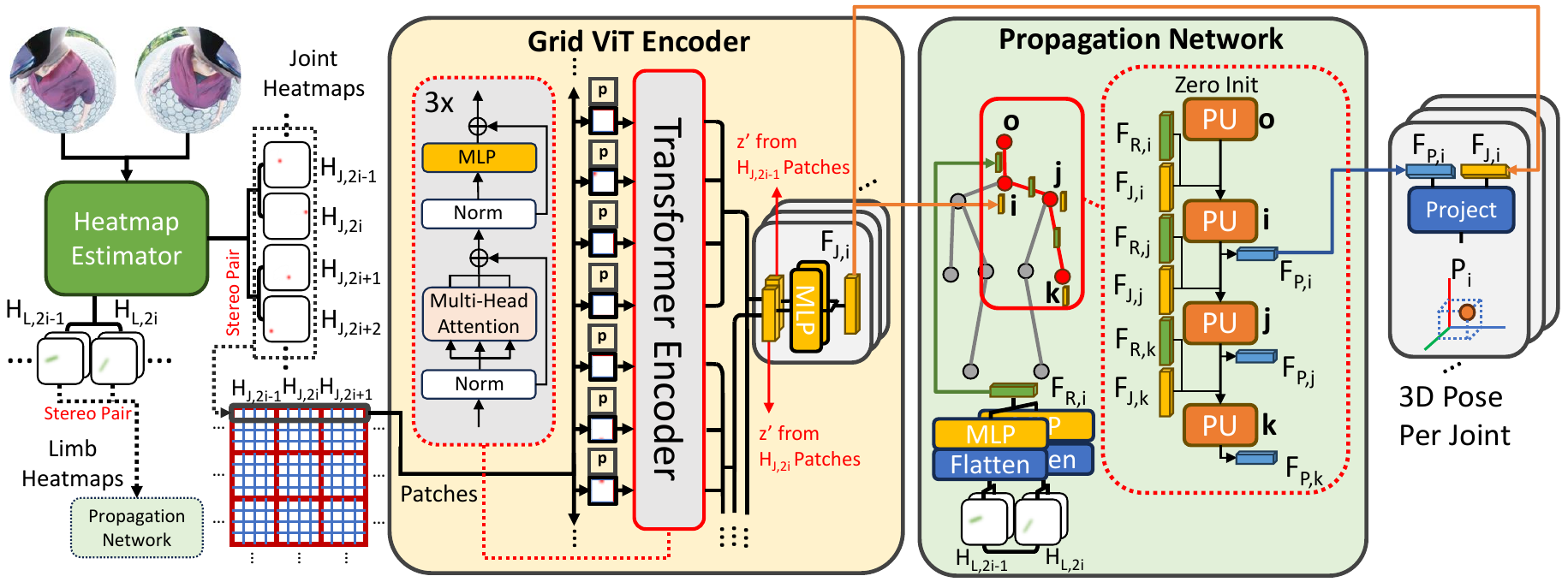}
    \end{subfigure}
    \caption{Overall network architecture of \pp{}. EgoTAP takes heatmaps from pre-trained heatmap estimators taking stereo input images and lifts the heatmaps to the 3D pose with the Grid ViT Encoder, Propagation Network, and finally, a projection layer.}
    \label{fig:architecture}
\end{figure*}

\noindent\textbf{Overall Architecture}. Fig.~\ref{fig:architecture} illustrates the comprehensive architecture of \pp{}. It comprises two essential components: the Grid ViT Heatmap Encoder and the Propagation Network. The Grid ViT Heatmap Encoder takes joint heatmaps as input and generates effective feature embeddings for each joint. The Propagation Network processes these embeddings with awareness of the skeletal structure to estimate the 3D pose accurately. Notably, the per-joint feature embedding is propagated through a skeletal hierarchy, represented as a tree structure with a root representing the head. In Fig.~\ref{fig:architecture}, a simplified skeleton is depicted, showcasing the propagation from the head to the hand, highlighted in red. The feature propagation utilizes the PU (Propagation Unit in Fig.~\ref{fig:propagation_cell}), which calculates joint states based on the parent joint's states along with other self-joint features. The hidden states of the last PU layer are concatenated with the joint features from the Grid ViT encoder and linearly projected to estimate the 3D pose of each joint.

\noindent\textbf{Input and Output}. Our method utilizes a pre-trained and frozen heatmap estimator that takes stereo RGB images $I \in \mathbb{R}^{2 \times 256 \times 256 \times 3}$ and estimates stereo heatmaps for $N_J$ joints $\mathbf{H_J} \in \mathbb{R}^{2N_J \times 64 \times 64}$ and $N_L$ limbs $\mathbf{H_L} \in \mathbb{R}^{2N_L \times 2 \times 64 \times 64}$. \pp{} takes the heatmaps and reconstructs the 3D pose $P \in \mathbb{R}^{N'_J \times 3}$ of $N'_J$ joints relative to the user's root defined in the dataset. Note that the number of estimation targets $N'_J$ can differ from the number of joints with heatmap $N_J$ depending on the dataset.

\noindent\textbf{Loss}. We use the Euclidean distance and the cosine similarity-based loss between the ground-truth pose and the estimated pose to train the Attention-Propagation network. The loss formulation is in the supplementary material.

\noindent\textbf{Heatmaps}. Two types of heatmaps for joints and limbs are used. We follow the standard definition of joint heatmap~\cite{conf/cvpr/TompsonGJLB15} where pixel values represent the probability that the joint is in that 2D coordinate. The limb heatmaps have two channels and are used to get relational features between two joints for the Propagation Network in Sec.~\ref{sec:propagation-network}. We use a limb heatmap suggested by Kang et al.~\cite{kang2023ego3dpose}, representing 3D information along with limb visibility as a line connecting joints. From the next section, we denote two types of heatmaps: \textit{joint heatmaps} and \textit{limb heatmaps}. We use a pre-trained ResNet-18~\cite{he2016residual} based U-Net~\cite{ronneberger2015convolutional} architecture with a shared weight for two input image encoders and shared decoder, suggested by Akada et al.~\cite{hakada2022unrealego} for heatmap estimation.

\subsection{Grid ViT Heatmap Encoder}
Our encoder, described in Fig.~\ref{fig:architecture}, combines all joint heatmaps into a large single grid image. The grid is split into patches, linearly projected to make the input embedding, and fed to a transformer~\cite{NIPS2017_3f5ee243} encoder architecture with multi-head attention. The transformer encoding process preserves the correspondence between a patch and the input feature embedding in the output. The output feature embeddings corresponding to individual input patches are concatenated and re-encoded to form a feature embedding vector for the heatmap.

Unlike the CNN encoder, where the communication occurs within the nearby pixels of different heatmaps, the Grid ViT Heatmap Encoder allows communication between heatmap patches that are far spatially. This allows features to be shared without downsampling, minimizing the loss of information. The efficiency of the encoder is demonstrated by the precisely reconstructed heatmaps from the embeddings in Fig.~\ref{fig:heatmap_reconstruction} and Table~\ref{table:effect_encoding}, and improved pose estimation accuracy.

To formulate the process, let $\{\mathbf{H_{J,i}} \in \mathbb{R}^{64 \times 64} | i = 1, 2, \ldots, 2N_J\}$ be sets of $2 \times N_J$ stereo joint heatmaps. Heatmaps are arranged into a single grid image. The image is subsequently split to total $4 \times 4 \times 2N_J$ patches $\{X_i \in \mathbb{R}^{16 \times 16} | i = 1, 2, \ldots, 32N_J\}$ where 16 patches corresponds to a heatmap. $X_{16(i-1)+1}$ to $X_{16i}$ corresponds to $i$-th heatmap for simplicity.

Each patch $X_i$ is then projected to an input embedding space $\mathbb{R}^{1024}$ with a learnable projection matrix $W_z \in \mathbb{R}^{1024 \times 256}$. Additionally, learnable positional encodings $\mathbf{p}_i \in \mathbb{R}^{1024}$ are added, resulting in the transformer input embedding $z_i$. The projected embedding with positional encoding for each patch is:
\begin{equation} z_i = W_z \cdot Flatten(X_i) + \mathbf{p}_i \end{equation}
$z = [z_1, z_2, \ldots, z_{32N_J}] $ is encoded by three ViT transformer encoder~\cite{dosovitskiy2020vit} layers with multi-head attention to output $z^\prime = [z^\prime_1, z^\prime_2, \ldots, z^\prime_{32N_J}]$. For the $j$-th heatmap, the corresponding output embeddings from 16 patches are concatenated to $\mathcal{Z}_j$ and then re-encoded to smaller dimensional feature embedding $k_j$ through multiple fully connected layers denoted as $E_K$. The process is formulated as follows:

\begin{equation}
z^\prime = \textit{TransformerEncoder}(z)
\end{equation}
\begin{equation} 
\mathcal{Z}_j = [z^\prime_{16(j-1)+1}, z^\prime_{16(j-1)+2}, \ldots, z^\prime_{16j}]
\end{equation}
\begin{equation} 
k_j = E_K(\mathcal{Z}_j)
\end{equation}

A joint feature $\mathbf{F_{J,i}} \in {R}^{256}$ that corresponds to a specific joint is obtained by concatenating the stereo heatmap features. Let's say $2i-1$ and $2i$-th heatmap correspond to $i$-th joint.

\begin{equation}
\mathbf{F_{J,i}} = [k_{2i-1}, k_{2i}], \text{ for } 1 \leq i \leq N_J
\end{equation}

\subsection{Propagation Network}
\label{sec:propagation-network}
\noindent\textbf{Propagation Process}. The Propagation Network estimates the joint positions using their parent joints' positions and the relationships between the joints. The Propagation Network is inspired by the stereo setup's capability to estimate 3D pose without the help of other joints and the general trend of higher visibility on joints closer to the camera in the egocentric setup. Sec.~\ref{sec:propagation-ablation} shows that the Propagation Network effectively takes advantage of accurate estimation of the parent joint with a Propagation Potential and Propagation Effect metric.

The Propagation Network comprises a relational feature encoder and the 2-layered PU that handles the propagation process. The relational feature encoder takes the estimated limb heatmaps to output the relational feature between joints. The PU handles the propagation process, which takes the parent states, relational and joint features of the child joint as input and generates the child joint's states. The states of joints are propagated through the tree hierarchy from the head directly attached to the camera to the extremities. During propagation, the reflection of the parent joint information is flexibly determined based on the certainty of the parent and child joint features by the PU.

We leverage the limb heatmaps with 3D information embedded with a trigonometric function of camera view angle~\cite{kang2023ego3dpose} to provide information about the connection between the parent and child joint. An encoder with fully connected layers $E_R$ encodes limb heatmaps $\mathbf{H_{L,i}} \in R^{2 \times 64 \times 64}$ into a limb feature. Stereo limb features are concatenated to form relational feature $\mathbf{F_R}$. Let's say $\mathbf{H_{L,2i-1}}$ and $\mathbf{H_{L,2i}}$ corresponds to a limb that connects the $i$-th joint and its parent. The process is:

\begin{equation}
\mathbf{F_{R,i}} = [E_L(\mathbf{H_{L,2i-1}}), E_L(\mathbf{H_{L,2i}})], \text{ for } 1 \leq i \leq N_L
\end{equation}

The Propagation Network consists of two layers of the Propagation Unit, described later. For a tree hierarchy where $parent(i)$ denotes a parent joint's index, and $\textit{PropagationNet}((H, C), R, J)$ denotes the Propagation Network, which takes hidden and cell states for two PU layers $H = [h_1, h_2]$, $C = [c_1, c_2]$, relational feature $R$ and joint feature $J$, the hidden and cell state for $i$-th joint $\mathbf{H}_i, \mathbf{C}_i$ is computed as follows:

\begin{equation}
\mathbf{S}_i = (\mathbf{H}_i, \mathbf{C}_i)
\end{equation}
\begin{equation}
\mathbf{H}_0 = \vec{0}, \mathbf{C}_0 = \vec{0}
\end{equation}
\begin{equation}
\mathbf{S}_i = \textit{PropagationNet}(\mathbf{S}_{\text{parent}(i)}, \mathbf{F_{J,i}}, \mathbf{F_{R,i}}), \text{ for } 1 \leq i \leq N_J
\end{equation}

The root joint head is indexed 0 and initialized with zero vector, as it is not visible from an egocentric view and, thus, does not have features. The $i$-th Propagated Feature $\mathbf{F_{P,i}} \in {R}^{256}$ is a hidden state from the second layer of the Propagation Network $\mathbf{h}_{2, i}$.

The output of the Propagation Network $\mathbf{F_{P,i}}$ and transformer output joint features $\mathbf{F_{J,i}}$ for each joint are concatenated and projected to estimate the 3D position of each joint.

\label{sec:propagation-unit}
\begin{figure}[pt]
    \centering
    \includegraphics[width=0.75\linewidth]{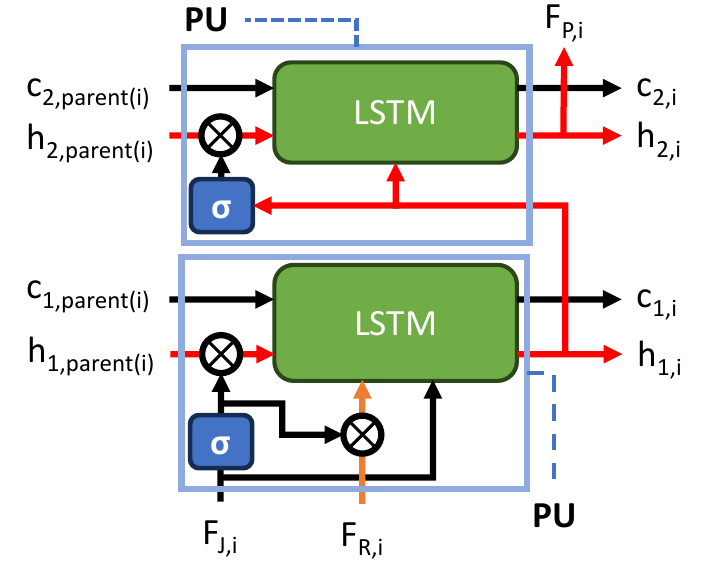}
    \caption{The Propagation Network with two layers of Propagation Unit.}
    \label{fig:propagation_cell}
\end{figure}

\noindent\textbf{Propagation Unit}. We devise a Propagation Unit inspired by the LSTM cell for the above propagation process. Fig.~\ref{fig:propagation_cell} shows the internal structure of the Propagation Unit. The Propagation Unit weights the parent's hidden state and the relational feature with the joint feature. The joint heatmap from stereo views can be sufficient for precise 3D estimation, and this weighting limits the role of the predictive estimation for obscure joints.

To formulate the Propagation Unit, we denote the weight matrix as $W$ and bias vectors as $b$. The symbol $\odot$ represents element-wise multiplication. The $+$ sign represents element-wise addition. $\sigma$ denotes the sigmoid activation.

\begin{equation}
f'_i = \sigma(W_{f'} \cdot \mathbf{F_{J,i}} + b_{f'})
\end{equation}
\begin{equation}
f''_i = \sigma(W_{f''} \cdot \mathbf{F_{J,i}} + b_{f''})
\end{equation}
\begin{equation}
h'_i = f'_i \odot h_{parent(i)}
\end{equation}
\begin{equation}
r'_i = f''_i \odot \mathbf{F_{R,i}}
\end{equation}
An additional forget gate is computed from the joint feature and is denoted as $f'_i$ and $f''_i$. The additional forget gate controls both the parent joint's hidden state and the relational feature between two joints, resulting in the modified hidden state $h'_i$ and the modified relational feature $r'_i$. Subsequently, these modified states and the joint feature treated as input are used in the standard LSTM architecture, weighted, and then applied non-linearity for the four gates: input, candidate cell state, forget, and output.

For the second layer of the Propagation Network, as there is only a hidden state from the previous layer without relational or joint feature distinction, the hidden state from the previous layer is used for forgetting the parent joint's hidden state in the current layer.
\section{Evaluation}
\subsection{Experiment Setup}
\subsubsection{Datasets}
\noindent\textbf{Overview}. We used two datasets: UnrealEgo~\cite{hakada2022unrealego} and EgoCap~\cite{rhodin2016egocap} for the 3D pose estimation in the stereo egocentric camera setup.
We conducted the within-dataset evaluation using each dataset's train and test set split since the egocentric datasets have significantly different setups and resulting views.

\noindent\textbf{UnrealEgo}. The UnrealEgo~\cite{hakada2022unrealego} is a synthetic dataset containing 450k frames with 17 characters. The dataset covers a variety of environments and motions that are challenging to capture in a real-world setup. There are a total of $16$ joints to estimate. The dataset defines the target local 3D pose in a pelvis-relative coordinate system, as opposed to the camera coordinate system in most datasets, and has a head pose to estimate. The pelvis and head do not have corresponding heatmaps and features. We added a learnable matrix for linear projection, taking all the final features $\mathbf{F_J}$ and $\mathbf{F_P}$ to estimate offset for all joints and head pose. We found that this simple change effectively deals with different pose definitions.

\noindent\textbf{EgoCap}. The EgoCap~\cite{rhodin2016egocap} dataset is captured with egocentric cameras attached at the end of the stick on the helmet. It comprises 35k frames for training from six subjects and 1k for testing from one subject with 3D pose annotation. Evaluation with this dataset showcases applicability in a real-world textured image. There are a total of $17$ joints to estimate.

\subsubsection{Baselines}
We experiment with three baseline stereo egocentric pose estimation methods: EgoGlass~\cite{zhao2021egoglass}, UnrealEgo~\cite{hakada2022unrealego}, and Ego3DPose~\cite{kang2023ego3dpose}. We use the official UnrealEgo~\cite{hakada2022unrealego} and Ego3DPose~\cite{kang2023ego3dpose} implementations. EgoGlass~\cite{zhao2021egoglass} implementation is taken from the latter as no official source code is provided. For the UnrealEgo~\cite{hakada2022unrealego} and Ego3DPose~\cite{kang2023ego3dpose}, we changed the embedding and pose decoder dimension, which gives higher estimation accuracy than their original setups. The change does not impact the EgoGlass~\cite{zhao2021egoglass}, possibly due to the joint training of the heatmap and pose estimator.

\subsubsection{Metrics}
The MPJPE and PA-MPJPE metrics are used. The MPJPE is a mean per joint position error in a 3D Euclidian distance. PA-MPJPE applies Procrustes analysis before computing the MPJPE to calculate transform-invariant positional error.

\subsection{Overall Performance}
\begin{figure}[pt]
    \begin{subfigure}{1.0\linewidth}
        \centering
        \includegraphics[width=\linewidth]{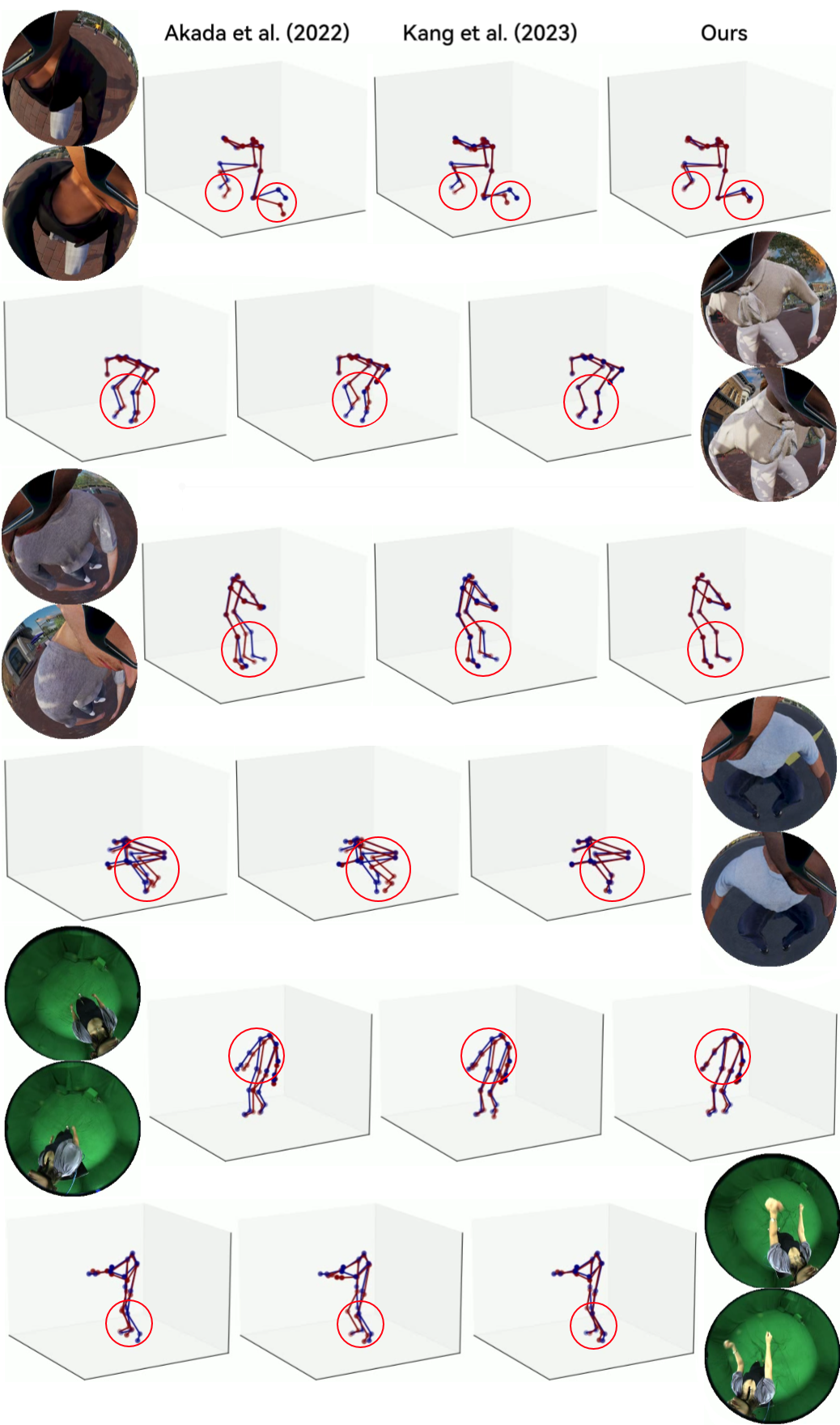}
    \end{subfigure}%
    \caption{Qualitative comparison of \pp{} with state-of-the-art stereo egocentric pose estimation methods. The blue is the ground truth, and the red is the estimated pose.}
    \label{fig:qualitative}
\end{figure}

\subsubsection{Qualatative Results}

Fig.~\ref{fig:qualitative} presents a qualitative comparison between our method and previous approaches on the UnrealEgo and EgoCap datasets. A more detailed qualitative comparison is available in the supplementary video. Our method demonstrates a significant improvement over baseline methods.

\subsubsection{Evaluation on UnrealEgo}
The second column of Table~\ref{table:unrealego_comparison} presents the quantitative evaluation results on UnrealEgo~\cite{hakada2022unrealego} using MPJPE and PA-MPJPE metrics. Our method demonstrates superior performance compared to state-of-the-art methods, achieving a 23.9\% reduction in MPJPE and a 17.7\% decrease in PA-MPJPE. These improvements extend across all 31 activity categories detailed in the supplementary material, covering a range of movements from common actions like sitting and standing to less frequent crawling and crouching and more complex motion categories, including sports.

Noteworthy improvements are observed across various categories, with the most substantial enhancement in the ``Crouching-Forward" category, boasting a 31.3\% reduction in MPJPE. Conversely, the smallest improvement is noted in the ``Crawling" activity, with an 8.8\% decrease in MPJPE. It's important to acknowledge that while our method relies on visual cues, the effectiveness varies based on the visibility of body parts. For instance, in activities like ``Crouching-Forward," where many body parts are partially visible, our method excels in improving accuracy. On the other hand, in activities like ``Crawling," where visible body features are significantly lacking, the challenge of enhancement is more pronounced.

\begin{table}\small
  \centering
  \begin{tabular}{@{}lccccc@{}}
  \toprule
    Method & UnrealEgo~\cite{hakada2022unrealego} & EgoCap~\cite{rhodin2016egocap} \\
    \midrule
    EgoGlass~\cite{zhao2021egoglass} & 81.55 (61.56) & 67.90 (-) \\
    UnrealEgo~\cite{hakada2022unrealego} & 63.53 (47.76) & 70.77 (52.91) \\
    Ego3DPose~\cite{kang2023ego3dpose} & 53.99 (43.02) & 69.45 (49.98) \\
    Ours & \textbf{41.06} (\textbf{35.39}) & \textbf{55.38} (\textbf{45.24}) \\
    \bottomrule
  \end{tabular}
  \caption{Evaluation results of state-of-the-art methods and ours on two datasets. The metric is MPJPE, and in the bracket is PA-MPJPE. The bold text indicates the best results.}
  \label{table:unrealego_comparison}
\end{table}

\subsubsection{Evaluation on EgoCap}
The third column of Table~\ref{table:unrealego_comparison} presents the quantitative results on the EgoCap dataset. Our method demonstrates significant outperformance, surpassing EgoGlass~\cite{zhao2021egoglass} by 22.6\% in MPJPE and Ego3DPose~\cite{kang2023ego3dpose} by 9.4\% in PA-MPJPE. For EgoGlass~\cite{zhao2021egoglass}, we report the MPJPE value from their paper, as they do not furnish official code or network details, and the available replication~\cite{kang2023ego3dpose} did not match the performance.

The relatively smaller improvement in PA-MPJPE, which discards the effect of the root's transform, could be attributed to prior methods estimating the full body pose as a whole. Consequently, they might capture the relative pose between joints while the estimation is globally biased. Nevertheless, when integrating the output camera coordinate system pose with the 6-DoF pose of VR and AR devices, precise pose estimation in the correct coordinate frame is crucial for accurate body tracking in the global coordinate system.

We observed that the estimated limb heatmaps in the EgoCap dataset exhibit lower accuracy than those in the UnrealEgo dataset, as illustrated in the supplementary material. This discrepancy could be attributed to the limited volume and the small number of subjects in the EgoCap dataset. Despite these challenges, our Attention-Propagation network effectively lifts the 3D pose from heatmaps. However, Ego3DPose~\cite{kang2023ego3dpose}, which utilizes limb heatmaps, did not perform well. This could be attributed to their explicit inference of orientation for each limb. The final decoder, which takes independent information as an output orientation, struggles with inaccurate information.

\subsection{Ablation Study}
We performed ablation studies to showcase the effectiveness of each network component, as summarized in Table~\ref{table:ablation}.

\begin{table}\small
  \centering
  \begin{tabular}{@{}lccccc@{}}
  \toprule
    Method & UnrealEgo~\cite{hakada2022unrealego} & EgoCap~\cite{rhodin2016egocap} \\
    \midrule
    \textbf{Heatmap Encoder} & & \\
    \midrule
    CNN & 63.53 (47.76) & 70.77 (52.91) \\
    Channel ViT & 61.62 (47.05) &  83.39 (56.29) \\ 
    Grid ViT & 49.03 (41.03) & 63.97 (53.17) \\ 
    \midrule
    \textbf{Propagation Network} & & \\
    \midrule
    Grid ViT + RF & 48.12 (40.79) & 63.09 (52.60) \\
    Grid ViT + LSTM & 49.43 (41.31) & 60.16 (49.18) \\ 
    Grid ViT + LSTM RF Alter & 44.97 (38.99) & 62.60 (50.78) \\ 
    Grid ViT + LSTM RF Concat & 44.77 (38.91) & 58.35 (47.06) \\ 
    \midrule
   \textbf{ Ours (Grid ViT + PU)} & \textbf{41.06} (\textbf{35.39}) & \textbf{55.38} (\textbf{45.24}) \\
    \bottomrule
  \end{tabular}
  \caption{Ablation results of our method for two main components on two datasets. The metric is MPJPE, and in the bracket is PA-MPJPE. The bold text for metrics indicates the best results.}
  \label{table:ablation}
\end{table}

\subsubsection{Grid ViT Heatmap Encoder}
\label{sec:heatmap-ablation}
\textbf{Pose Estimation:} We assess the impact of the Grid ViT Heatmap Encoder. ``CNN" presents the results from UnrealEgo~\cite{hakada2022unrealego}, utilizing a CNN. ``Channel ViT" showcases the outcomes with a typical encoder with ViT, where heatmaps are concatenated along the channel axis before being split into patches, resulting in feature embeddings that do not align with the heatmaps. Simply adopting transformers~\cite{NIPS2017_3f5ee243} yields minimal improvement, i.e., a 3\% reduction in MPJPE, compared to the CNN-based lifting for the UnrealEgo~\cite{hakada2022unrealego} baseline and dataset. However, this approach significantly degrades performance on EgoCap~\cite{rhodin2016egocap}. This observation underscores the importance of addressing the correspondence between feature embedding and heatmaps in the pose estimation process.

\begin{table}\small
  \centering
  \begin{tabular}{@{}lccccc@{}}
    \toprule
    Heatmap Reconstruction Error & $10^{-4}$/Pixel \\
    \midrule
    Zeros & 5.45 \\
    CNN Encoder & 4.84 \\
    Grid ViT Heatmap Encoder & 1.68 \\
    \bottomrule
  \end{tabular}
  \caption{Reconstruction mean square error of the heatmaps from the features encoded with a different frozen encoder architecture, experimented in the UnrealEgo~\cite{hakada2022unrealego} dataset.}
  \label{table:effect_encoding}
\end{table}

\textbf{Heatmap Reconstruction:} We conducted experiments to evaluate the heatmap encoder's efficiency in encoding heatmap features. A simple decoder is appended to our encoder and baseline encoders to achieve this. The decoder is trained to reconstruct the estimated heatmaps from the feature embedding. Table~\ref{table:effect_encoding} presents the reconstruction error of the heatmap in the test set. The ``Zeros" row provides the error for a zero-only output for comparison. The results demonstrate that the Grid ViT Heatmap Encoder effectively extracts heatmap features, evidenced by the reconstructed fine details of the heatmap in Fig.~\ref{fig:heatmap_reconstruction}. In contrast, the heatmaps were not recoverable from features encoded by CNN, highlighting its inefficiency.

\subsubsection{Propagation Network}
\label{sec:propagation-ablation}
\textbf{Pose Estimation:} We investigate if including relational features alone can significantly enhance accuracy through ``+ RF" when incorporated with our Grid ViT encoder. The relational features are concatenated to the joint features for the final projection layer without the involvement of a propagation network. This approach demonstrates marginal impact or even degrades the estimation accuracy. Additionally, we analyze the effect of the Propagation Network with LSTM~\cite{LSTM1997}. In the case of ``+ LSTM," only joint features are utilized in the propagation, yielding a marginal effect.

Additional experiments investigate the impact of the Propagation Network without PU, denoted as ``+ LSTM RF Alter" and ``+ LSTM RF Concat." Relational and joint features are alternately taken in the former, and the propagation feature is output in the joint feature step. The latter takes both as a concatenated vector. Both methods demonstrate improvements, with the latter achieving an 8.7\% and 8.8\% reduction in MPJPE for two datasets compared to the Grid ViT Heatmap Encoder-only approach. The final model, incorporating PU, maximizes the potential of the Propagation Network, showcasing a 16.3\% and 13.4\% improvement in MPJPE for the two datasets. This highlights the significance of balancing the role of predictive estimation using parent joints and direct estimation using self-joint features.

\begin{figure}[pt]
    \centering
    \begin{subfigure}{0.49\linewidth}
        \centering
        \includegraphics[width=\linewidth]{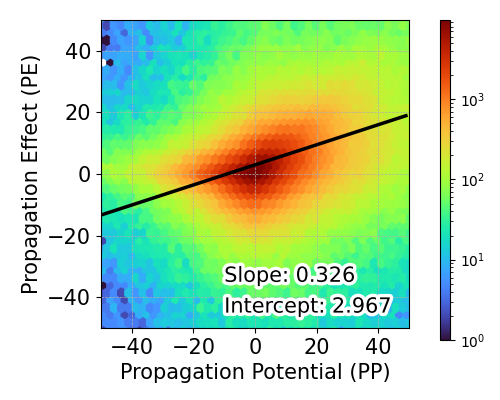}
        \caption{UnrealEgo~\cite{hakada2022unrealego}}
    \end{subfigure}%
    \hfill 
    \begin{subfigure}{0.49\linewidth}
        \centering
        \includegraphics[width=\linewidth]{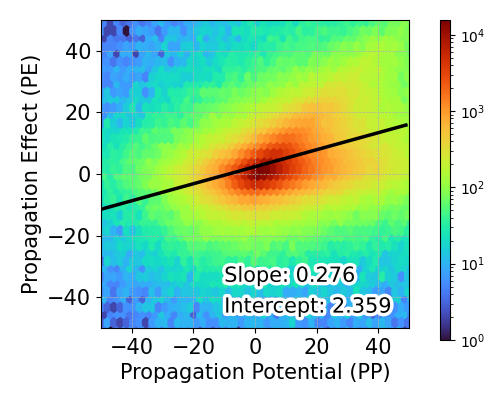}
        \caption{UnrealEgo (Camera-relative)}
    \end{subfigure}%
    
    \begin{subfigure}{\linewidth}
        \centering
        \includegraphics[width=0.49\linewidth]{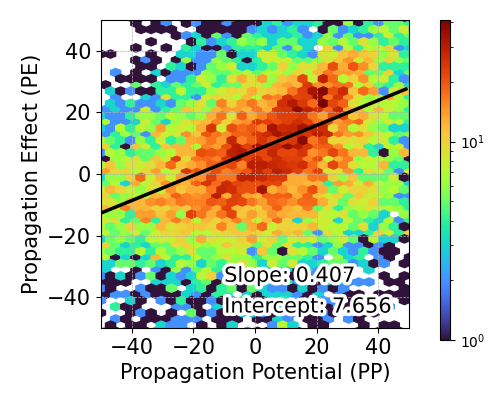}
        \caption{EgoCap~\cite{rhodin2016egocap}}
    \end{subfigure}%
    
    \caption{Hexagonal-grid density plot of the Propagation Potential and the Propagation Effect(mm) in our evaluation datasets. The dark line shows linear regression results.}
    \label{fig:propagation_analysis}
\end{figure}

\textbf{Propagation Potential and Effect:} 
The Propagation Network leverages more evident parent joint features to improve the child joint's pose estimation. The hexagonal-grid density plot in Fig.~\ref{fig:propagation_analysis} illustrates its impact quantitatively. The $x$-axis represents the Propagation Potential ($\mathbf{PP}$). $\mathbf{PP}$ approximates the upper bound of the improvement using the parent's feature, with a difference between the parent and child joint's pose estimation error. On the $y$-axis, the Propagation Effect ($\mathbf{PE}$) is the improvement of the child joint's pose error by the Propagation Network. Using $\Delta$ to denote the pose estimation error, subscripts to denote joints, and superscripts to denote the model ($\mathbf{NP}$ without propagation, $\mathbf{P}$ with propagation), we define these metrics as follows.

\begin{equation}
\mathbf{PP} = \Delta_{\text{child}}^{\mathbf{NP}} - \Delta_{\text{parent}}^{\mathbf{NP}}
\end{equation}
\begin{equation}
\mathbf{PE} = \Delta_{\text{child}}^{\mathbf{NP}} - \Delta_{\text{child}}^{\mathbf{P}}
\end{equation}

For all datasets, linear regression reveals a positive relationship between $\mathbf{PP}$ and $\mathbf{PE}$ with a p-value of the null hypothesis $<10^{-3}$, indicating that the Propagation Network is more effective when the parent joint has a more precise estimation, aligning with expectations. The average $\mathbf{PP}$ and $\mathbf{PE}$ were $16.97$ and $8.50$ for the UnrealEgo dataset~\cite{hakada2022unrealego} and $4.32$ and $9.39$ for the EgoCap~\cite{rhodin2016egocap} dataset. The UnrealEgo~\cite{hakada2022unrealego} dataset exhibits higher potential due to the cameras closer to the head, unlike cameras around 20cm away from the head in the EgoCap dataset~\cite{rhodin2016egocap}.

The effect is more pronounced for the UnrealEgo~\cite{hakada2022unrealego} dataset when the 3D pose is estimated in camera-relative coordinates. This eliminates the global offset (pelvis pose) bias from per-joint improvement. Fig.~\ref{fig:propagation_analysis} (b), exhibits trends where $\mathbf{PE}$ is similar to $\mathbf{PP}$ or close to zero. When the $\mathbf{PE}$ is similar to $\mathbf{PP}$, the child joint's pose error is improved close to the parent joint's error. The effect of the Propagation Network is near the upper bound ($\mathbf{PP}$). The propagation cannot improve the child joint's pose error in some cases, possibly due to the occlusion of limbs. Such cases exhibit near zero $\mathbf{PE}$. $66.07$\% of $\mathbf{PE}$ and $75.62$\% of $\mathbf{PP}$ in the samples are positive, and $54.16$\% of samples lie in the first quadrant. The average positive $\mathbf{PE}$ is $10.75$, while the average negative $\mathbf{PE}$ is only $-0.51$, demonstrating that many joints significantly benefit from the propagation.
\section{Conclusion}
In this study, we introduce a novel heatmap-to-3D lifting method tailored for the stereo egocentric setup, employing a transformer for efficient feature embedding and an attention-driven Propagation Network focused on evident features. We demonstrate effective heatmap feature extraction through the Grid ViT Heatmap Encoder, employing patch-wise communication with self-attention to preserve correspondence between the heatmap and the feature embedding. The Propagation Network utilizes visual cues from the proximate parent joint, leveraging joint relational information to predictively estimate less visible child joint poses. Our experiments highlight significant advancements over state-of-the-art stereo egocentric pose estimation methods, underscoring the efficacy of our proposed approach.

\section*{Acknowledgement}
This work was supported by the National Research Foundation of Korea(NRF) grant funded by the Korea government(MIST) (No. 2022R1A2C3008495). This work was supported by the National Research Foundation of Korea(NRF) grant funded by the Korea government(MSIT) (No.RS-2023-00218601).

\appendix
\clearpage
\setcounter{page}{1}
\maketitlesupplementary
\section{Overview}
The supplementary material contains the following:

\begin{itemize}
\item{Dataset Processing}
\item{Implementation}
\item{Training}
\item{Experiment}
\item{Example Figure}
\item{Limitations and Future Works}
\end{itemize}

\section{Dataset Processing}
We explain the details of the train and test dataset we used in this section. Our method requires a 2D and 3D pose annotation and stereo input images. The 2D annotation is necessary for generating the heatmaps.

\subsection{UnrealEgo}
We utilize the full dataset, including metadata files and preprocessed pickles. The public Ego3DPose~\cite{kang2023ego3dpose} code loads metadata and pickles. Their code adds 2D and 3D pose data in the camera coordinate system and their limb heatmap representation in the pickle files. Our method uses these final pickles.

\subsection{EgoCap}
We used publicly available 2D pose annotation on the train set. Additionally, we got the full ground truth 3D pose for the train set of the EgoCap~\cite{rhodin2016egocap} dataset from the authors. In the fisheye views of the dataset, images are projected only in the circular area due to strong distortion. Thus, the original images contain areas that do not have real views. Following the Kang et al.~\cite{kang2023ego3dpose}, we cropped the image horizontally into a square area centered at the x-axis focal center ($\mathbf{fx}$) provided in the dataset calibration data. We resized the images to 256 by 256 images to fit our model.

The dataset has a train set and 2D and 3D validation sets. The 3D validation set contains a ground truth 3D pose and is used for testing. The 2D validation dataset provides the 2D annotation for the images in the 3D validation sets from a subject labeled 7. The 3D pose is converted from a $mm$ to a $cm$ unit to scale the pose loss in accordance with the UnrealEgo dataset.

\section{Implementation}
\begin{figure}[pt]
    \centering
    \includegraphics[width=0.25\linewidth]{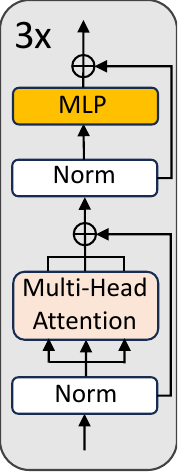}
    \caption{The ViT encoder architecture.}
    \label{fig:transformer_encoder_architecture}
\end{figure}

\subsection{Grid ViT Heatmap Encoder}
The $64 \times 64$ sized heatmaps are put into one image with resolution $384 \times 384$. The image comprises $36$ areas as a $6 \times 6$ grid. The number of joint heatmaps is $30$ for the UnrealEgo~\cite{hakada2022unrealego} dataset and $34$ for the EgoCap~\cite{rhodin2016egocap}. The heatmaps fill in the grid in order. The areas that do not correspond to any heatmap are masked in the ViT encoder module and don't impact the output.

We adopt the ViT encoder~\cite{dosovitskiy2020vit} architecture. Our implementation adopts the public Transformers~\cite{wolf-etal-2020-transformers} module \textbf{ViTModel} class for the PyTorch~\cite{NEURIPS2019_9015}. We removed the $[CLS]$ token since we are not using the module for a classification task. Doing so improves pose estimation accuracy empirically. The module follows the standard ViT~\cite{dosovitskiy2020vit} encoder architecture shown in Fig.~\ref{fig:transformer_encoder_architecture} that takes the input embedding $z$ and outputs feature embedding $z^\prime$.

The ViT encoder takes embeddings of size $1024$ per each of $32N_J$ patches, $z = [z_1, z_2, \ldots, z_{32N_J}]$. The multi-head attention layer has $8$ heads. The intermediate layer size of the MLP is $4096$. The Grid ViT Heatmap Encoder uses three ViT encoder layers. It outputs a total $16384$ size of the embedding vector from $16$ patches for each heatmap. The embedding vector is then compressed with MLP denoted $E_\mathbf{K}$ in the paper. The MLP has ReLU~\cite{agarap2018learning} non-linearity for the intermediate layers. The MLP's hidden sizes of the first two layers are $2048$ and $512$, and the last layer outputs a final embedding of size $128$.

\subsection{Propagation Network}
In an extension of the typical LSTM~\cite{LSTM1997}, the Propagation Unit's relational features, joint features, hidden and cell states, and gate outputs all have the same size. We chose $256$ for the size.

\subsubsection{Limb Heatmap Encoder}
The limb heatmap encoder $E_R$ extracts relational features. The encoder consists of three layers with the same structure as the final MLP layers of the Grid ViT Heatmap Encoder, with only an input size difference. The input two-channeled limb heatmap~\cite{kang2023ego3dpose} has $2 \times 64 \times 64$ size. The encoder takes it after flattening it. The encoder consists of three fully connected layers, the first two layers with $2048$ and $512$ output size, with the ReLU~\cite{agarap2018learning} activation, and the final layer outputs the embedding with a size $128$.

\subsubsection{Second Layer of the PU}
\label{sec:2nd_pu}
The second layer of PU does not take distinct relational and joint features. It takes the parent joint's second layer cell and hidden state with the first layer's hidden state of the joint. Since hidden states from different layers are used in this section, let's denote the $n$-th layer hidden states of $i$-th joint $h_{n,i}$. The additional forget gate in the second layer $g_i$ controls the parent joint's second PU layer's hidden state, resulting in the modified hidden state $h^\prime_{2,i}$. This is formulated as follows:

\begin{equation}
g_i = \sigma(W_{g} \cdot h_{1,i} + b_{g})
\end{equation}
\begin{equation}
h^\prime_{2,i} = g_i \odot h_{2,{parent(i)}}
\end{equation}

The modified parent hidden state and the joint's first layer hidden state are input for the inner LSTM~\cite{LSTM1997}.

\subsubsection{Internal LSTM of the PU}
We explain the formulation of the LSTM~\cite{LSTM1997} inside the PU in more detail here.

\noindent\textbf{Formulation of typical LSTM}. The LSTM is formulated as follows, where $h_{i-1}$ denotes the hidden state of the previous step, $c_{i-1}$ denotes the cell state of the previous step, and $x_i$ denotes the input. Here, $W$ and $b$ denote weights and biases for each gate. The symbol $\odot$ represents element-wise multiplication, and the $+$ sign represents element-wise addition. $\tanh$ and $\sigma$ denote the hyperbolic tangent and sigmoid activation.

\begin{equation}
f_i = \sigma(W_f \cdot [h_{i-1}, x_i] + b_f)
\end{equation}
\begin{equation}
i_i = \sigma(W_i \cdot [h_{i-1}, x_i] + b_i)
\end{equation}
\begin{equation}
o_i = \sigma(W_o \cdot [h_{i-1}, x_i] + b_o)
\end{equation}
\begin{equation}
\tilde{c}_i = \tanh(W_c \cdot [h_{i-1}, x_i] + b_c)
\end{equation}
\begin{equation}
c_i = f_i \odot c_{i-1} + i_i \odot \tilde{c}_i
\end{equation}
\begin{equation}
h_i = o_i \odot \tanh(c_i)
\end{equation}
The $f_i$, $i_i$, and $o_i$ are forget, input, and output gates. $\tilde{c}_i$ denotes the candidate cell value. $h_i$ and $c_i$ are the final hidden and cell state for step $i$.

\noindent\textbf{Formulation of internal LSTM}. Unlike the LSTM taking the cell and hidden state, the internal LSTM of the first PU layer takes three states in addition to input joint features. The three states are the modified parent's hidden state $h^\prime_i$, the modified relational feature of the joint $r^\prime_i$, and the cell state of the parent $c_{parent(i)}$. The input is joint features $\mathbf{F_{J,i}}$.

This section explains the first and second layers together; thus, we denote the $n$-th layer of $i$-th joint with a $n,i$ subscript, as in $h_{n,i}$ for the hidden state.
In the computation of the forget, input, and output gates and the candidate cell value, a concatenated vector of the modified parent's hidden state and relational features, and the joint features $[h^\prime_{1,i}, r^\prime_{1,i}, \mathbf{F_{J,i}}]$ replaces $[h_{i-1}, x_i]$. 

\begin{equation}
f_{1,i} = \sigma(W_{1,f} \cdot [h^\prime_{1,i}, r^\prime_i, \mathbf{F_{J,i}}] + b_{1,f})
\end{equation}
\begin{equation}
i_{1,i} = \sigma(W_{1,i} \cdot [h^\prime_{1,i}, r^\prime_i, \mathbf{F_{J,i}}] + b_{1,i})
\end{equation}
\begin{equation}
o_{1,i} = \sigma(W_{1,o} \cdot [h^\prime_{1,i}, r^\prime_i, \mathbf{F_{J,i}}] + b_{1,o})
\end{equation}
\begin{equation}
\tilde{c}_{1,i} = \tanh(W_{1,c} \cdot [h^\prime_{1,i}, r^\prime_i, \mathbf{F_{J,i}}] + b_{1,c})
\end{equation}

For the second layer, the modified second layer parent hidden state $h^\prime_{2,i}$ from the Sec.~\ref{sec:2nd_pu} takes the place of $h_{i-1}$. The previous layer's hidden state $h_{1,i}$ replaces input $x_i$, analogous to the standard multi-layered LSTM.

\begin{equation}
f_{2,i} = \sigma(W_{2,f} \cdot [h^\prime_{2,i}, h_{1,i}] + b_{2,f})
\end{equation}
\begin{equation}
i_{2,i} = \sigma(W_{2,i} \cdot [h^\prime_{2,i}, h_{1,i}] + b_{2,i})
\end{equation}
\begin{equation}
o_{2,i} = \sigma(W_{2,o} \cdot [h^\prime_{2,i}, h_{1,i}] + b_{2,o})
\end{equation}
\begin{equation}
\tilde{c}_{2,i} = \tanh(W_{2,c} \cdot [h^\prime_{2,i}, h_{1,i}] + b_{2,c})
\end{equation}

The Propagation Unit takes features from the parent joint, not the previous index. In the computation of the final cell and hidden state, both layers of PU take $c_{n,parent(i)}$ instead of $c_{i-1}$ in the formula. The hidden state is computed in the same way.
\begin{equation}
c_{n,i} = f_{n,i} \odot c_{n,parent(i)} + i_{n,i} \odot \tilde{c}_{n,i}
\end{equation}
\begin{equation}
h_{n,i} = o_{n,i} \odot \tanh(c_{n,i})
\end{equation}

\section{Training}
\subsection{Hardware Setup}
We trained and tested our method on a server with NVIDIA RTX A6000 GPU and AMD EPYC 7313 16-Core Processor CPU.

\subsection{Heatmap Estimator}
The heatmap estimator is trained using UnrealEgo~\cite{hakada2022unrealego} code and their scripts for the UnrealEgo dataset. The default configuration utilizes Adam~\cite{KingBa15} optimizer with a learning rate $10^{-3}$. They train the network for $10$ epochs, the later $5$ epochs with linear decay, with batch size $16$.
For the EgoCap dataset, we trained the heatmap estimators for $30$ epochs with the same setup. Linear decay is used for the last $15$ epochs proportionally.

When hasty convergence, where all heatmap values converge to 0, is detected, the training is automatically restarted, following the protocol of Kang et al.~\cite{kang2023ego3dpose}.

\subsection{\pp{} Network}
The network is trained with the AdamW~\cite{loshchilov2018decoupled} optimizer with pretrained and frozen heatmap estimator weight. The learning rate of $10^{-3}$ is used with 16 epochs with a cosine annealing scheduler, with batch size $32$. Early epochs use a linear warmup, one epoch for the UnrealEgo~\cite{hakada2022unrealego}, and two epochs for the EgoCap~\cite{rhodin2016egocap} dataset.

\subsection{Loss}
\subsubsection{\pp{} Network}
\pp{} has two loss terms: a pose error loss (i.e., joints' average Euclidean distance) and cosine-similarity loss~\cite{tome2019xr} that focuses on estimating the correct 3D orientation for each limb.

The pose error loss is defined as follows. Given two 3D joint poses: the predicted pose \( \mathbf{p\prime}_i \) and the ground truth pose \( \mathbf{p}_i \), for \( i = 1, \ldots, J \), where \( J \) is the total number of joints:

\begin{equation}
{L_{p}} = \frac{1}{J} \sum_{i=1}^{J} \| \mathbf{p\prime}_i - \mathbf{p}_i \|_2
\end{equation}

The cosine similarity loss is then defined as follows. A limb pose vector for a particular joint is obtained by subtracting the pose of its parent joint from its pose, i.e., \( \mathbf{v}_i = \mathbf{p}_i - \mathbf{p}_{\text{parent}(i)} \) and \( \mathbf{v\prime}_i = \mathbf{p\prime}_i - \mathbf{p\prime}_{\text{parent}(i)} \) for the ground truth and predicted poses, respectively. Given these vectors, the cosine similarity between two limb pose vectors is calculated using their inner product:

\begin{equation}
{L_{c}} =  \frac{1}{J-1} \sum_{i=2}^{J} \| \frac{\mathbf{v}_i \cdot \mathbf{v\prime}_i}{\| \mathbf{v}_i \|_2 \| \mathbf{v\prime}_i \|_2}
\end{equation}

Note that the root joint is ignored since it does not have a parent joint.

The final loss term is a weighted sum of two losses, where we choose $w_p = 0.1$ and $w_c = -0.01$. The cosine similarity loss weight has a negative sign because higher cosine similarity indicates a more accurate pose.

\begin{equation}
L = w_p L_p + w_c L_c
\end{equation}

\subsubsection{Heatmap Reconstruction Network for Ablation}
We adopted the heatmap decoder proposed by Tome et al.~\cite{tome2019xr} for the heatmap reconstruction in the ablation study of the Grid ViT Heatmp Encoder.
The network with only mean squared loss struggles from early convergence to outputting near-zero valued heatmaps. The problem is more severe than the heatmap estimator network since the reconstruction network does not contain specialized architecture like the U-Net~\cite{ronneberger2015convolutional}, which helps the heatmap generation through the multi-resolution features.

Thus, additional loss to match the heatmap's minimum and maximum values is added if the mean squared loss is higher than the threshold to prevent the network training in the ablation studies from converging to outputting only zeros. We set the threshold to a value empirically found sufficient to ensure avoidance of the zero-only convergence. The loss term guides the network to output a peak in the heatmap, as joint heatmaps do.

The heatmap reconstruction's target is to minimize the mean squared loss between the predicted heatmap $H$, and reconstructed heatmap $H\prime$. This is computed as follows:

\begin{equation}
L_r = \frac{1}{N} \sum_{i=1}^{N} (H_i - H'_i)^2
\end{equation}

The min-max loss applies only if the mean squared loss is higher than threshold $\theta$. The threshold is $5.5 * 10^{-4}$. 
It is computed as follows:

\begin{equation}
L_{min} = \frac{1}{N} \sum_{i=1}^{N} |\min(H_i) - \min(H'_i)|
\end{equation}

\begin{equation}
L_{max} = \frac{1}{N} \sum_{i=1}^{N} |\max(H_i) - \max(H'_i)|
\end{equation}

\begin{equation}
L_m = 
\left\{
\begin{aligned}
& L_{min} + L_{max} & \text{if } L_r > \theta \\
&0 & \text{otherwise}
\end{aligned}
\right.
\end{equation}

The weight for the reconstruction $w_r$ is set to $1$ and the weight for the min-max penalty $w_m$ is set to $10^{-3}$, resulting in the total loss:

\begin{equation}
L = w_r \cdot (L_r + w_m \cdot L_m)
\end{equation}

The loss term does not impact the final result once the network avoids the early convergence and stabilizes. The zero output convergence is still observed for the CNN encoder embedding, so the training was restarted until it did not converge to output only zeros. Such an additional loss term is necessary to get meaningful non-zero reconstruction from the output of the CNN heatmap encoder, showing the difficulty of heatmap information recovery from its embeddings.

\section{Experiment}

\begin{table*}[!]
\tiny
\centering
\begin{tabular}{@{}lcccccccc@{}}
    \toprule
    \multicolumn{9}{c}{MPJPE (PA-MPJPE)} \\
    \midrule
    Method & Jumping & Falling Down & Exercising & Pulling & Singing & Rolling & Crawling & Laying \\
    \midrule
    EgoGlass~\cite{zhao2021egoglass} & 78.93(63.85) & 123.80(92.71) & 94.21(69.85) & 79.41(55.41) & 68.16(50.25) & 100.53(87.26) & 173.69(111.51) & 106.41(86.42) \\
    UnrealEgo~\cite{hakada2022unrealego} & 61.66(49.46) & 108.73(78.02) & 77.14(58.87) & 57.01(43.51) & 52.61(37.58) & 73.38(64.56) & 162.90(102.15) & 82.60(67.47) \\
    Ego3DPose~\cite{kang2023ego3dpose} & 52.12(43.29) & 86.08(71.72) & 67.52(56.39) & 48.92(37.02) & 43.86(34.54) & 74.24(64.81) & 138.47(92.93) & 78.13(67.23) \\
    \textbf{Ours} & \textbf{43.05(37.31)} & \textbf{75.77(63.48)} & \textbf{52.76(46.21)} & \textbf{34.45(26.82)} & \textbf{33.96(29.10)} & \textbf{52.24(47.58)} & \textbf{126.23(91.46)} & \textbf{66.38(59.56)} \\
    \midrule
    Method & Sitting on the Ground & Crouching & Crouching and Turning & Crouching to Standing & Crouching-Forward & Crouching-Backward & Crouching-Sideways & Standing-Whole Body \\
    \midrule
    EgoGlass & 204.35(147.99) & 121.76(100.12) & 130.24(104.28) & 84.31(59.67) & 82.84(66.06) & 90.36(76.83) & 101.37(78.81) & 69.78(52.59) \\
    UnrealEgo & 190.26(144.27) & 96.69(79.29) & 116.59(99.94) & 66.20(44.92) & 56.10(46.62) & 62.54(46.21) & 72.35(55.87) & 52.91(39.14) \\
    Ego3DPose & 143.44(122.93) & 82.01(67.94) & 104.24(83.98) & 58.74(41.34) & 48.60(38.81) & 47.36(36.05) & 57.85(48.83) & 45.69(35.47) \\
    \textbf{Ours} & \textbf{121.24(103.27)} & \textbf{67.27(60.07)} & \textbf{89.97(70.78)} & \textbf{41.91(28.68)} & \textbf{33.47(29.52)} & \textbf{34.08(28.30)} & \textbf{40.21(38.51)} & \textbf{32.77(28.27)} \\
    \midrule
    Method & Standing-Upper Body & Standing-Turning & Standing to Crouching & Standing-Forward & Standing-Backward & Standing-Sideways & Dancing & Boxing \\
    \midrule
    EgoGlass & 69.24(49.36) & 77.77(60.27) & 83.86(81.65) & 76.75(63.23) & 78.40(59.83) & 82.71(66.46) & 82.84(65.59) & 66.98(49.13) \\
    UnrealEgo & 50.97(34.86) & 60.42(46.27) & 48.09(40.5) & 56.23(47.90) & 57.14(44.90) & 63.10(50.86) & 64.73(51.79) & 52.13(38.36) \\
    Ego3DPose & 44.14(32.99) & 51.45(41.91) & 55.66(45.08) & 49.48(44.65) & 45.35(36.52) & 52.25(44.97) & 55.30(46.47) & 41.55(32.14) \\
    \textbf{Ours} & \textbf{31.29(25.51)} & \textbf{41.24(35.55)} & \textbf{37.07(30.28)} & \textbf{39.64(38.06)} & \textbf{32.43(30.25)} & \textbf{39.34(37.94)} & \textbf{42.14(38.39)} & \textbf{29.97(25.69)} \\
    \midrule
    Method & Wrestling & Soccer & Baseball & Basketball & American Football & Golf & & \\
    \midrule
    EgoGlass & 84.23(62.81) & 81.57(60.59) & 76.20(56.11) & 78.33(57.30) & 102.54(84.03) & 69.69(48.15) & & \\
    UnrealEgo & 67.85(52.73) & 67.09(51.43) & 62.15(48.60) & 64.73(47.79) & 89.57(68.49) & 55.87(40.34) & & \\
    Ego3DPose & 57.96(45.94) & 59.56(45.23) & 56.21(42.17) & 56.02(41.94) & 77.89(62.56) & 48.10(36.01) & & \\
    \textbf{Ours} & \textbf{44.15(39.38)} & \textbf{48.27(38.56)} & \textbf{44.83(35.92)} & \textbf{45.19(36.78)} & \textbf{65.30(54.41)} & \textbf{38.97(31.25)} & & \\
\bottomrule
\end{tabular}
\caption{Quantitative evaluation results on the \textbf{UnrealEgo} dataset per category.}
\label{table:unrealego_categorical}
\end{table*}

\section{Categorical Evaluation on the UnrealEgo dataset}
Table \ref{table:unrealego_categorical} categorically shows the result on the UnrealEgo~\cite{hakada2022unrealego} dataset. Metrics from all three baseline methods, EgoGlass~\cite{zhao2021egoglass}, UnrealEgo~\cite{hakada2022unrealego}, and Ego3DPose~\cite{kang2023ego3dpose} are shown with our method. The MPJPE values are outside the bracket, and the PA-MPJPE values are inside the bracket.

\begin{figure}
    \begin{subfigure}{1.0\linewidth}
        \centering
        \begin{subfigure}{1.0\linewidth}
            \centering
            \includegraphics[width=\linewidth]{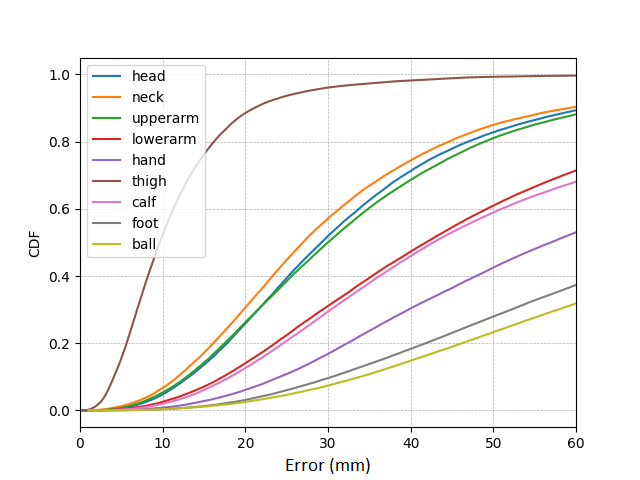}
        \end{subfigure}
        \caption{UnrealEgo~\cite{hakada2022unrealego}}
        \begin{subfigure}{1.0\linewidth}
            \centering
            \includegraphics[width=\linewidth]{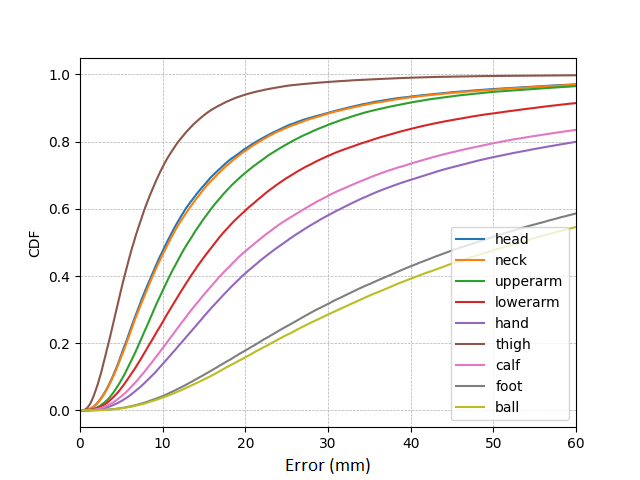}
        \end{subfigure}
        \caption{Ours}
    \end{subfigure}
    \caption{CDF of errors for each joint in the UnrealEgo~\cite{hakada2022unrealego} dataset with their method.}\label{fig:joint_distribution}
\end{figure}

\subsection{Per Joint Error Distribution}
Fig.~\ref{fig:joint_distribution} shows the CDF (Cumulative Distribution Function) of the error of each joint. Two results show one from the UnrealEgo~\cite{hakada2022unrealego} method as an example of the baseline in the introduction and one for our method, both evaluated on the UnrealEgo dataset.

The thigh is directly attached to the pelvis, which is the origin of the local pose definition in the dataset. Thus, for all methods, the thigh has the lowest errors among all joints. Lower body joints, calf, foot, and balls generally have significantly higher errors than other joints, except for the hands, which have larger errors than the calf. The error of the hand and the lower arm gets much closer to the upper arm in our method, showing the benefit of the propagation.

\begin{figure}
    \begin{subfigure}{0.48\linewidth}
        \centering
        \begin{subfigure}{0.48\linewidth}
            \centering
            \includegraphics[width=\linewidth]{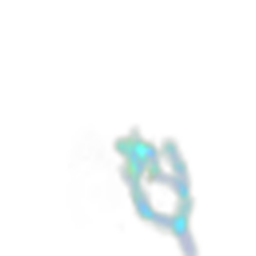}
        \end{subfigure}%
        \begin{subfigure}{0.48\linewidth}
            \centering
            \includegraphics[width=\linewidth]{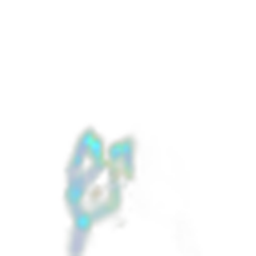}
        \end{subfigure}
        \caption{EgoCap~\cite{rhodin2016egocap} dataset}
    \end{subfigure}
    \begin{subfigure}{0.48\linewidth}
        \centering
        \begin{subfigure}{0.48\linewidth}
            \includegraphics[width=\linewidth]{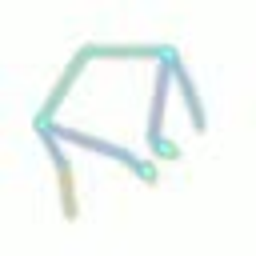}
        \end{subfigure}%
        \begin{subfigure}{0.48\linewidth}
            \centering
            \includegraphics[width=\linewidth]{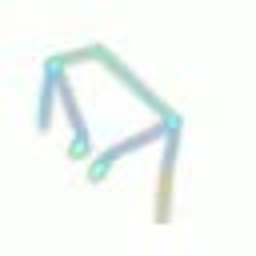}
        \end{subfigure}
        \caption{UnrealEgo~\cite{hakada2022unrealego} dataset}
    \end{subfigure}
    \caption{Estimated limb heatmaps on the test set of the \textbf{EgoCap} (Left) and \textbf{UnrealEgo} (Right).}
    \label{fig:heatmaps_comparison}
\end{figure}

\begin{table}\small
  \centering
  \begin{tabular}{@{}lccccc@{}}
  \toprule
     & UnrealEgo~\cite{hakada2022unrealego} & EgoCap~\cite{rhodin2016egocap} \\
    \midrule
    Estimated Heatmap & 41.06 & 55.38 \\
    Ground Truth Heatmap & 6.63 & 26.63 \\
    \bottomrule
  \end{tabular}
  \caption{Comparison of pose estimation error (MPJPE) of our method, with estimated and ground truth heatmaps provided as input. Columns indicates two datasets.}
  \label{table:gt_heatmap_comparison}
\end{table}
\subsection{Impact of the Heatmap Estimation Accuracy}
We experimented with our architecture's performance when the ground truth heatmaps were provided instead of the estimated heatmaps. Table~\ref{table:gt_heatmap_comparison} reveals that the limited 2D pose information from the view is a key bottleneck for egocentric pose estimation. The full 2D pose provided by the ground truth heatmap reduces error significantly. Despite the better view provided by the camera attached far from the head, the EgoCap has a higher estimation error with ground truth heatmaps. A relatively small dataset volume for training can also be a bottleneck.

\subsection{Impact of ViT Backbone Size}
Experiments revealed that the bottleneck of the pose estimation accuracy is not in the computational capacity of the backbone. We experimented with up to 12 layers of the ViT encoders and 8 times larger feature sizes in the ViT encoder. No notable improvement was observed compared to the smaller backbone we chose. The UnrealEgo~\cite{hakada2022unrealego} shows consistent experimental results that the larger ResNet backbones do not improve the pose estimation accuracy.

\section{Example Figure}
\subsection{Limb Heatmaps}
The main text mentions that the limb heatmap estimation is less accurate on the EgoCap~\cite{rhodin2016egocap}. The heatmap visualization in Fig.~\ref{fig:heatmaps_comparison} shows noisy lines for limbs.

\section{Limitations and Future Works}
\pp{} is limited to a single frame input and relies fully on visual cues. The result with the \pp{} on motions with severe occlusion, such as ``Crawling'' and ``Sitting on the Ground'', has very high error compared to other motion categories as shown in Table~\ref{table:unrealego_categorical}. Unlike many recently proposed general pose estimation methods, the egocentric setup's exploration of utilizing the temporal context is limited. For the egocentric view with a limited view, the invisible joints' pose can benefit significantly from the temporal context. For one example in the egocentric setup, Wang. et al.~\cite{Wang_2021_ICCV} applied temporal optimization using a variational autoencoder for improved pose estimation in the global coordinate.

The method's applicability can further be tested on monocular and different potential egocentric camera setups. The Propagation Network is based on the stereo setup, which provides sufficient information for a 3D pose when the joint is visible from both views. Thus, the propagation scheme helps child joint pose estimation. While the 3D pose estimation from the single heatmap is not feasible in the monocular setup, pose space is highly constrained, and our method can also be applicable potentially with modification. 

The Propagation Network applies to an egocentric view with a specific characteristic. The method itself lacks dynamicity like the GCN-based method~\cite{GCN2021Pose}, which would make it applicable to many different situations. The tree hierarchy assumption still holds for arbitrary root joints in the skeletal hierarchy, giving room for more dynamicity. Applying such a tree hierarchy-based network has the potential for a specific joint-related situation, such as collision. Such application remains a future work.

{
    \small
    \bibliographystyle{ieeenat_fullname}
    \bibliography{main}
}


\end{document}